\documentclass{ieeeaccess}
\usepackage{cite}
\usepackage{amsmath,amssymb,amsfonts}
\usepackage{algorithmic}
\usepackage{algorithm}
\usepackage{textcomp}
\usepackage{graphicx}
\usepackage{lipsum} 
\usepackage{calc}   
\usepackage{subcaption}  
\usepackage{grffile}
\usepackage{caption}        
\usepackage{multirow}       
\usepackage{booktabs}       
\usepackage[colorlinks=true, linkcolor=blue, citecolor=blue, urlcolor=blue]{hyperref}
\def\BibTeX{{\rm B\kern-.05em{\sc i\kern-.025em b}\kern-.08em
    T\kern-.1667em\lower.7ex\hbox{E}\kern-.125emX}}

\makeatletter
\def\@doi{}
\makeatother

\begin{document}

\makeatletter
\def\ps@accessfooter{%
  \def\@oddfoot{\hfill\thepage}%
  \def\@evenfoot{\hfill\thepage}%
  \let\@oddhead\@empty
  \let\@evenhead\@empty}
\makeatother

\pagestyle{accessfooter}

\title{Dynamic Meta-Ensemble Framework for Efficient and Accurate Deep Learning in Plant Leaf Disease Detection on Resource-Constrained Edge Devices}
\author{\uppercase{Weloday Fikadu Moges}\authorrefmark{1*}, 
\uppercase{Jianmei Su}\authorrefmark{1},\uppercase{Amin Waqas}\authorrefmark{2}}
\address[1*]{South West University Science and Technology, School of Computer Science and Technology, Mianyang, China}
\address[1]{South West University Science and Technology, School of Computer Science and Technology, Mianyang, China }
\address[2]{South West University Science and Technology, School of Control and Information Engineering Technology, Mianyang, China }

\corresp{ Weloday Fikadu Moges (deyu@mails.swust.edu.cn)}

\begin{abstract}
Deploying deep learning models for plant disease detection on edge devices such as IoT sensors, smartphones, and embedded systems is severely constrained by limited computational resources and energy budgets. To address this challenge, we introduce a novel Dynamic Meta-Ensemble Framework (DMEF) for high accuracy plant disease diagnosis under resource constraints. DMEF employs an adaptive weighting mechanism that dynamically combines the predictions of three lightweight convolution neural networks (\mbox{MobileNetV2}, \mbox{NASNetMobile}, and \mbox{InceptionV3}) by optimizing a trade-off between accuracy improvements ($\Delta$Acc) and computational efficiency (model size). During training, the ensemble weights are updated iteratively, favoring models exhibiting high performance and low complexity. Extensive experiments on benchmark datasets for potato and maize diseases demonstrate state-of-the-art classification accuracies of 99.53\% and 96.61\%, respectively, surpassing standalone models and static ensembles by 2.1--6.3\%. With computationally efficient inference latency (\textless 75 ms ) and a compact footprint (\textless 1 million parameters), DMEF shows strong potential for edge-based agricultural monitoring, suggesting viability for scalable crop disease management. This bridges the gap between high-accuracy AI and practical field applications.
\end{abstract}


\titlepgskip=-15pt

\maketitle
\thispagestyle{empty} 
\section{Introduction}
\label{sec:introduction}
The timely detection of plant leaf diseases represents a critical challenge in precision agriculture, serving as the first line of defense against catastrophic crop losses \cite{ref17}. In particular, in economically vital crops such as potatoes and maize, undiagnosed pathologies can trigger production declines that exceed 20-40\% \cite{ref18}, exacerbating global food insecurity. Although the conventional diagnosis of plant diseases predominantly depends on visual inspection performed by trained agronomists, a method that is labor-intensive, time-consuming, and prone to human error, it often results in inconsistent and inaccurate disease identification. Such limitations hinder timely intervention and can exacerbate crop losses, highlighting the need for more reliable automated detection techniques\cite{ref23}.

\indent Contemporary deep learning architectures, particularly ResNet \cite{ref1}, Inception \cite{ref2}, and VGG \cite{ref20} variants have demonstrated remarkable success in plant disease classification, achieving $>$95\% accuracy in controlled laboratory settings. However, their translation to practical edge deployment faces three fundamental barriers: (1) the computational intractability of billion-parameter models on IoT-grade hardware \cite{ref24}, (2) memory constraints in embedded systems \cite{ref24}, and (3) unsustainable power consumption for continuous field operation \cite{ref25}. These limitations make state-of-the-art models challenging for real-world agricultural deployment.
 
\indent Emerging compact architectures (MobileNet \cite{ref4}, EfficientNet \cite{ref5}, SqueezeNet \cite{ref6}) to reconcile this tension through architectural pruning and quantization, but have drawbacks. For instance, SqueezeNet shows very fast inference and small model size but suffers from poor accuracy on complex datasets. At the same time, EfficientNetV2 achieves high accuracy but has larger size and slower inference, limiting its use in resource-constrained settings\cite{ref27}. Recent research also explores low-cost ensemble methods to reduce overhead but notes that achieving the same performance as classical ensembles often still requires multiple forward passes or complex model architectures, thus maintaining non-trivial computation costs\cite{ref26}.

\indent In this paper, we propose a Dynamic Meta-Ensemble Framework for efficient and accurate plant leaf disease detection on resource-constrained edge devices. The framework dynamically adjusts the contribution of each model in an ensemble based on two key factors: (1) the accuracy improvement during training and (2) the model size during inference. By prioritizing accuracy during training and emphasizing smaller, efficient models during inference,the framework ensures optimal performance on edge devices. Our approach is designed to balance accuracy and computational efficiency, making it suitable for real-time disease detection in agricultural applications.

\subsection{Motivation and Contribution}
\subsubsection{Motivation}
Food security is increasingly threatened by crop diseases, yet traditional detection methods are slow and unreliable. While AI solutions offer promise, they often fail to balance accuracy with the computational constraints of edge devices used in real-world farming. Our work addresses this gap by introducing a dynamic lightweight AI system that delivers lab-grade disease detection to farmers via affordable, edge-compatible devices.
\subsubsection{Contribution}
We propose a dynamic meta ensemble framework that adapts to agricultural challenges by: (1) introducing a novel weighting mechanism to balance accuracy and efficiency for edge deployment; (2) designing an architecture optimized for real-time processing; (3) demonstrating consistent, superior performance across various crops. This work bridges the gap between high-accuracy AI and practical field applications, setting a new benchmark for intelligent crop disease monitoring.
\vspace{2pt}
\section{RELATED WORKS}
Deep learning has revolutionized computer vision, achieving state-of-the-art performance in image classification tasks. In agriculture, deep learning models have been widely adopted for plant disease detection. Early work by Mohanty et al. \cite{ref16} demonstrated the potential of convolutional neural networks (CNNs), achieving 99.35\% accuracy on lab-captured images of 26 diseases across 14 plant species. However, these models struggled in real-world field conditions due to variability in lighting, occlusion, and background noise. Subsequent studies, such as Sladojevic et al. \cite{ref21}, achieved high accuracy (up to 96.3\%) in classifying 13 leaf diseases across 5 plant species using deep neural networks; however, their results were based on a relatively small and limited dataset, which constrains the generalizability of their approach to broader, real-world scenarios. Ferentinos \cite{ref12} further advanced the field by developing a deep CNN-based system that achieved up to 99.53\% accuracy in classifying 58 different plant disease classes using the PlantVillage dataset. While this work demonstrated state-of-the-art performance, the computational demands of such deep learning models remain a significant barrier for real-time or edge deployment, as these models typically require substantial processing power and memory. These studies highlight the potential of deep learning for automating plant disease detection. However, most existing approaches rely on large, high-accuracy models such as ResNet and Inception, which are computationally intensive and unsuitable for deployment on resource-constrained edge devices. Although these models achieve high accuracy, their computational cost limits their practicality in real-world agricultural applications.

Recent research has introduced innovative architectures such as depthwise CNNs with squeeze-and-excitation (SE) blocks (EfficientNet \cite{ref5}) and residual connections (Xception \cite{ref14}), achieving high accuracy and enhancing computational efficiency and real-time applicability. Transfer learning approaches, notably those leveraging Xception \cite{ref14} and advanced data augmentation, have further improved robustness and generalization across multiple plant species and disease types \cite{ref15,ref9}. Despite these advances, challenges persist, including data imbalance, model interpretability, and generalization to novel diseases in diverse field conditions \cite{ref9,ref13}.

State-of-the-art architectures like ResNet \cite{ref1} and Inception \cite{ref2} deliver high accuracy but are computationally intensive (e.g., VGG-16 requires 15.5 GFLOPs) \cite{ref3}. Lightweight alternatives such as MobileNet \cite{ref4} and EfficientNet \cite{ref5} reduce parameters and computational cost, with models like SqueezeNet achieving AlexNet-level accuracy with 50× fewer parameters \cite{ref6}. However, these lightweight models often sacrifice performance in complex field conditions. For example, in plant disease detection tasks, MobileNetV2 achieved up to 92.8\% accuracy, which is typically lower than ResNet-50, which can exceed 98\% accuracy on the same datasets \cite{ref8}.

Ensemble techniques such as voting and averaging combine the strengths of multiple deep learning models to enhance the accuracy and robustness of plant leaf disease detection in a variety of crops and datasets, it has been extensively shown that these methods perform better than single-model solutions, yielding gains in precision, recall, F1-score, and overall classification accuracy \cite{ref7} but despite accuracy gains, ensemble methods typically impose significantly higher computational costs. This is because they require running several models in parallel, each performing feature extraction and prediction, which increases memory usage, processing time, and energy consumption\cite{ref19}. Static ensembles lack adaptability to dynamic field conditions and fail to optimize the accuracy-efficiency tradeoff. Recent work has explored attention mechanisms (e.g., ECA modules \cite{ref10}) and pruning strategies (e.g., CACPNET \cite{ref11}) to balance accuracy and resource use, but these often require manual tuning and do not fully address scalability or deployment constraints \cite{ref12,ref13}.

Current approaches face a trilemma: high-accuracy models are resource-heavy, lightweight models lack robustness, and static ensembles are inflexible. Emerging directions include integrating explainable AI for model interpretability, leveraging IoT and edge computing for real-time deployment, and employing advanced data augmentation and transfer learning for better generalization \cite{ref12,ref13}. The development of dynamic, adaptive frameworks—such as the proposed DMEF—aims to bridge these gaps, offering scalable, efficient, and accurate solutions for plant disease diagnosis in practical agricultural environments.

Our framework addresses these limitations by integrating meta-learning principles with dynamic weight updates based on both accuracy improvements and model size. By explicitly quantifying complexity through parameter counts and updating weights epoch-wise, we ensure the ensemble remains efficient without sacrificing accuracy—a critical advantage for real-world applications like plant disease identification.
\vspace{2pt} 
\section{ PROPOSED DMEF METHODOLOGY}
\begin{figure}[h]
    \centering
    \includegraphics[width=0.9\textwidth]{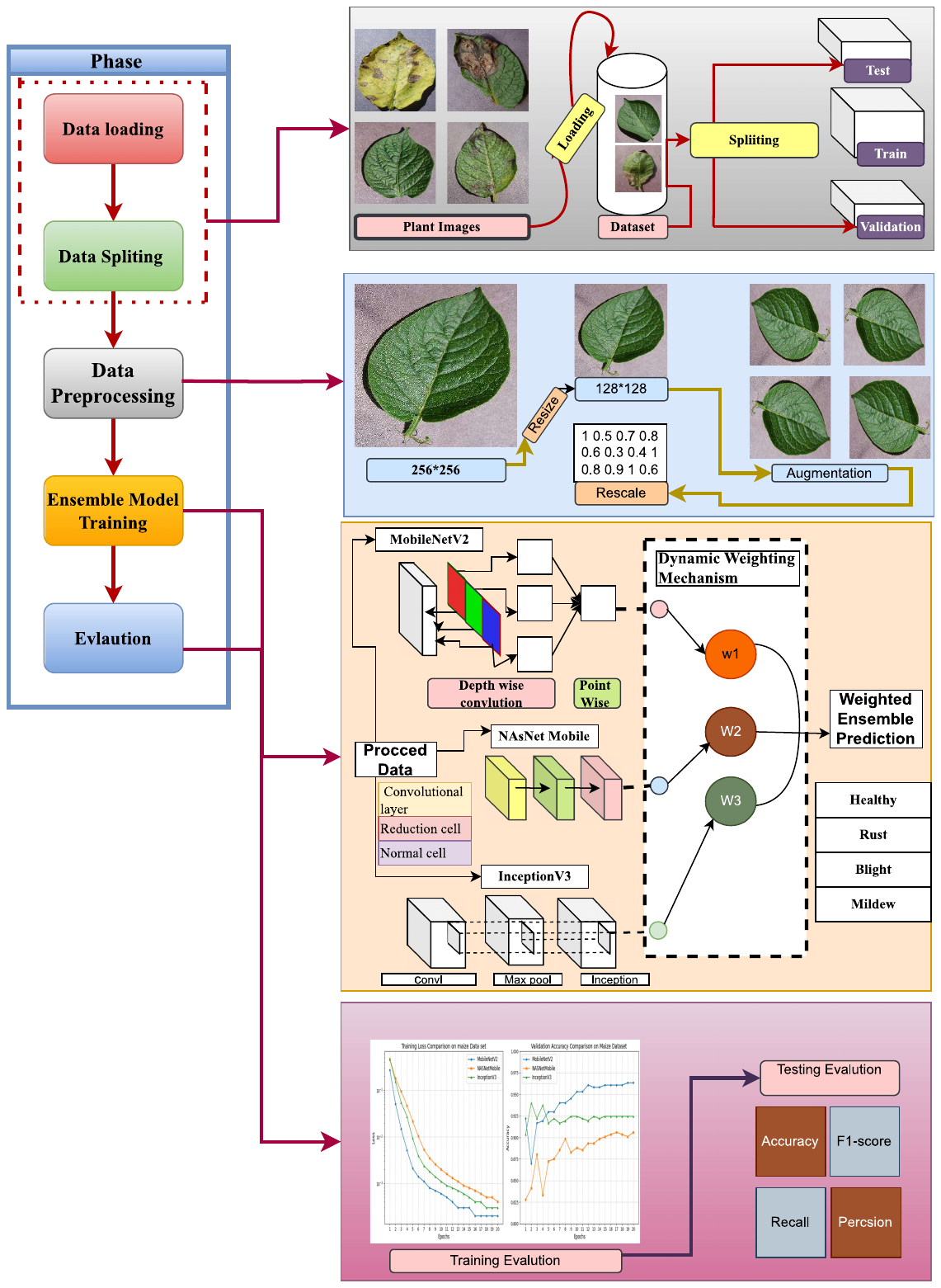} 
    \caption{ Dynamic Meta-Ensemble Framework for Plant Disease Classification}
    \label{Fig 1}
\end{figure}

The proposed Dynamic Meta-Ensemble Framework (DMEF) methodology begins with comprehensive data loading, splitting, and preprocessing, which includes data augmentation techniques. Following this, employs a diverse set of lightweight base models with varying architectures and complexities. During the training phase, the ensemble dynamically updates the weights of each model by monitoring their accuracy improvements and adjusting a balancing parameter that controls the influence of accuracy versus model size. This adaptive process ensures that models with recent performance gains receive higher weights, while also considering computational cost. At inference, the ensemble aggregates weighted predictions from all base models to produce the final classification output.
The proposed methodology is illustrated in \autoref{Fig 1}. To improve clarity, the methodology is divided into four parts: (I) data preprocessing and data augmentation, (II) base model architecture and model description, (III) dynamic meta-ensemble framework and training algorithm, and (IV) inference, evaluation, and symbol definition.
\vspace{0.25cm}
\subsection{Data Preprocessing and Data Augmentation}
Effective data preprocessing and augmentation are critical for training robust and generalizable deep learning models. In this study, we utilize the publicly available Plant Village (PV) dataset, which comprises a diverse and representative collection of RGB images depicting various plant diseases in 
multiple crop species. Each image in the dataset is resized to a standardized resolution of $128\times128$ pixels. This step not only ensures compatibility with the input requirements of modern convolutional neural network (CNN) architectures but also helps reduce computational overhead during training and inference.

To further harmonize the data, pixel values are normalized to the $0-1$ range, thereby accelerating the model convergence and mitigating the adverse effects of illumination variability across samples. The dataset is subsequently partitioned into three mutually exclusive subsets using a stratified sampling strategy to preserve the original class distribution.Specifically, $80\%$ of the data is allocated for training, $10\%$ for validation, and the remaining $10\%$ for testing. The training set is utilized for model learning and parameter optimization, the validation set is employed for hyperparameter tuning and early stopping to prevent overfitting, and the test set is reserved for the final, unbiased evaluation of the framework's performance.  

 A critical aspect of the preprocessing pipeline is the application of advanced data augmentation techniques, which are exclusively performed on the training data, the whole techniques as Table~\ref{tab:data}. These augmentations include random horizontal and vertical flips, rotations at varying angles, zoom transformations, and contrast adjustments. Such operations are designed to simulate the natural variability, encountered in real-world agricultural environments, including changes in leaf orientation, scale, and lighting conditions. 

 \vspace{0.25cm}
 \begin{table}[ht]
\centering
\caption{Data augmentation layers and their effects}
\begin{tabular}{l l l}
\hline
\textbf{Layer} & \textbf{Parameter} & \textbf{Effect} \\ \hline
RandomFlip & horizontal\_and\_vertical& Flips images\\ 
RandomRotation & 0.2 & Rotates images by ±20° \\ 
RandomZoom  & 0.2  & Zooms in/out(±20°scale) \\ 
RandomContrast & 0.2& Adjusts contrast((±20°).\\ \hline
\end{tabular}
\label{tab:data}
\end{table}

By artificially expanding the diversity of the training data, these augmentations play a pivotal role in enhancing the model's ability to generalize to unseen samples and reduce the risk of overfitting. This careful and systematic approach to data preparation ensures that the resulting models are both accurate and resilient, capable of delivering reliable predictions across a wide range of plant disease scenarios.

\vspace{0.5cm}
\subsection{Base Model Architecture and Model Description}
To capitalize on the rich and diverse training and validation datasets prepared through rigorous preprocessing and augmentation, adopts a transfer learning strategy utilizing three state-of-the-art convolutional neural network (CNN) architectures: MobileNetV2, NASNetMobile, and InceptionV3.The selection of these models is motivated by their complementary strengths in balancing predictive accuracy, computational efficiency, and adaptability to resource-constrained
environments. By leveraging these diverse architectures, the framework aims to maximize generalization and robustness in plant disease detection across a wide range of real-world scenarios.
\onecolumn
\twocolumn

MobileNetV2 is chosen for its lightweight design and exceptional efficiency, which utilizes depth-wise separable convolutions and inverted residual blocks with linear bottlenecks \cite{ref22}, and is included as one of the base models due to its reduced parameter count and computational requirements. In our implementation, MobileNetV2 comprises 417,284 trainable parameters, a testament to its compactness and suitability for real-time applications in intelligent agriculture and field diagnostics.

NASNetMobile, developed through Neural Architecture Search (NAS), is included for its ability to automatically optimize both accuracy and efficiency \cite{ref35}. Its modular design, composed of normal and reduction cells, allows the network to adaptively learn hierarchical representations while maintaining a low computational footprint. With 174,420 trainable parameters, NASNetMobile provides a compelling balance between model complexity and inference speed, making it ideal for scenarios where both accuracy and efficiency are paramount.

InceptionV3 is selected for its proven high accuracy and sophisticated architectural innovations, such as inception modules \cite{ref36} that process input data through multiple parallel convolutional filters of varying sizes, enabling the extraction of rich multi-scale features. It further reduces computational complexity by factorizing larger convolutions and employs auxiliary classifiers to improve gradient flow during training. InceptionV3, with 402,308 trainable parameters, serves as a robust feature extractor and has consistently demonstrated superior performance in plant disease classification tasks.

All three models are initialized with pretrained ImageNet weights, providing a strong foundation of generic visual features. To adapt these models to the specific nuances of the Plant Village dataset, the top classification layers are removed and replaced with a global average pooling layer followed by a dense softmax output layer corresponding to the number of disease classes. During training, only the later layers of each model are unfrozen: three layers for MobileNetV2, twenty layers for NASNetMobile, and thirteen layers for InceptionV3. This selective fine-tuning approach aligns with best practices in transfer learning. The rationale is that the early layers of convolutional neural networks capture universal low-level features such as edges, textures, and simple shapes, which are generally transferable across different visual domains. In contrast, the later layers encode higher-level, task-specific representations. By unfreezing only the last layers, the models adapt to the unique characteristics of plant disease imagery without overwriting the valuable general features learned from large-scale datasets. This strategy not only accelerates convergence but also mitigates the risk of overfitting.
\subsection{Dynamic Meta-Ensemble Ensemble Framework and Training Algorithm}
\begin{figure}[h]
    \centering
    \includegraphics[width=0.5\textwidth]{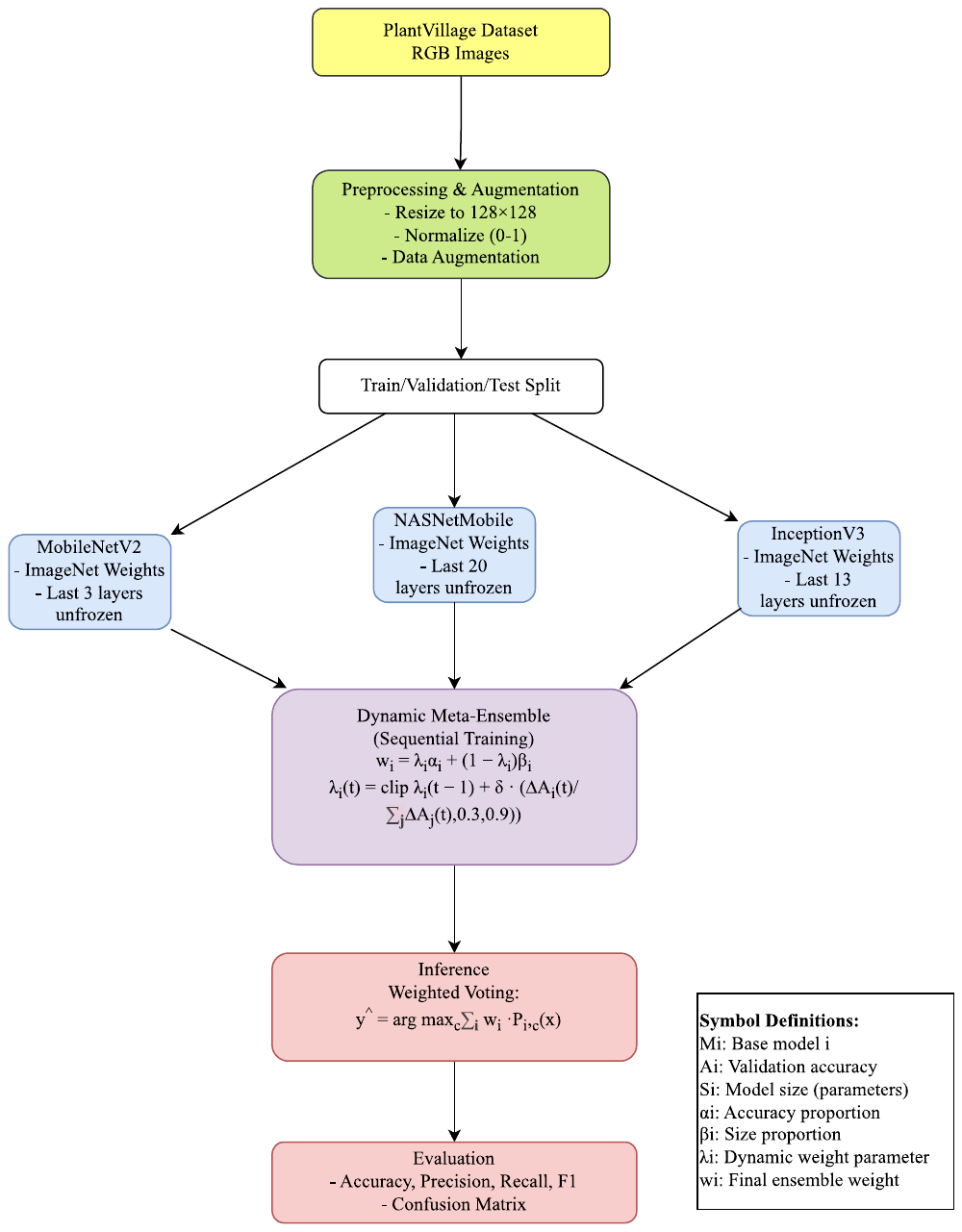} 
    \caption{ Flow chart of the Dynamic Meta-Ensemble Framework}
    \label{Fig 2}
\end{figure}

Following the careful preprocessing of the PlantVillage dataset and the fine-tuning of MobileNetV2, NASNetMobile, and InceptionV3, as detailed in the previous section, the complete workflow—encompassing these preprocessing and fine-tuning steps as well as subsequent stages—is depicted in \autoref{Fig 2}. Proceeding from these initial steps, the methodology advances to its core innovation: a dynamic meta-ensemble framework. This framework is designed to enhance classification performance by adaptively combining the strengths of multiple lightweight base classifiers. It employs a principled weighting mechanism that effectively balances predictive accuracy and model complexity.

A key aspect of this methodology is the implementation of a dynamic meta-ensemble strategy. During training, each model’s performance is evaluated at every epoch, and the corresponding weights are iteratively updated based on improvements in accuracy, the relative size of each model, and their contribution to overall performance. Model size is quantified by the total number of parameters, while the accuracy proportion reflects each model’s contribution to improved classification results. By dynamically adjusting the weights in this manner, the ensemble emphasizes models that balance accuracy and efficiency, resulting in a robust and effective classification system.
\subsubsection{Framework Overview}
Consider an ensemble composed of $n$ base classifiers $M_1, M_2, \ldots, M_n$, each trained independently on the prepared dataset. The core idea is to assign a dynamic weight $w_i$ to each classifier $M_i$ that reflects its current competence and computational cost, enabling the ensemble to emphasize models that are both accurate and efficient. Formally, the weight $w_i$ for the $i$-th classifier is defined as a convex combination of two normalized metrics: the accuracy proportion $\alpha_i$ and the model size proportion $\beta_i$:
\begin{equation}
w_i = \lambda_i \alpha_i + (1 - \lambda_i) \beta_i \tag{1}
\end{equation}
Here, the accuracy proportion $\alpha_i$ is computed as
\begin{equation}
\alpha_i = \frac{A_i}{\sum_{j=1}^n A_j} \tag{2}
\end{equation}
where $A_i$ denotes the validation accuracy of classifier $M_i$. This normalization ensures that the relative predictive strength of each model is considered in the ensemble weighting. Conversely, the model size proportion $\beta_i$ quantifies the relative complexity of $M_i$ by normalizing its number of trainable parameters $S_i$:
\begin{equation}
\beta_i = \frac{S_i}{\sum_{j=1}^n S_j} \tag{3}
\end{equation}
\subsubsection{Model Size Calculation and Importance}
The model size $S_i$ is calculated by aggregating the parameters across all layers of the network:
\begin{equation}
S_i = \sum_{l=1}^{L_i} \prod_{k=1}^{K_l} d_{l,k} \tag{4}
\end{equation}
where $L_i$ is the number of layers in $M_i$, $K_l$ is the dimensionality of the weight tensor at layer $l$, and $d_{l,k}$ represents the size of the $k$-th dimension in that tensor. This formulation captures the total number of trainable parameters, serving as a proxy for the computational cost and memory footprint of the model.

Calculating the number of trainable parameters is crucial for several reasons: it quantifies the model’s complexity and learning capacity, helps assess the risk of overfitting, and determines the computational resources required for training and inference. By incorporating $S_i$ into the ensemble weighting, the framework promotes a balance between predictive accuracy and computational efficiency, which is essential for deploying models in resource-constrained environments.
\subsubsection{ Dynamic Weight Update}

To dynamically balance the trade-off between accuracy and complexity, the parameter $\lambda_i$ is iteratively updated during training based on the relative accuracy improvement $\Delta A_i(t)$ observed at epoch $t$:
\begin{equation}
\lambda_i(t) = \text{clip}\left(\lambda_i(t-1) + \delta \cdot \frac{\Delta A_i(t)}{\sum_{j=1}^n \Delta A_j(t)}, \lambda_{\min}, \lambda_{\max}\right) \tag{5}
\end{equation}
where $\delta$ is a learning rate controlling the update magnitude, and the clipping function restricts $\lambda_i$ within predefined bounds $[\lambda_{\min}, \lambda_{\max}]$ (typically $[0.3, 0.9]$) to maintain stability. This adaptive update mechanism allows the ensemble to progressively favor classifiers demonstrating recent accuracy gains while preventing overemphasis on any single model.

The dynamic meta-ensemble framework adaptively combines predictions from the three base models using a weighting scheme that balances model accuracy and efficiency. For each training epoch, the accuracy of each model on the training set is computed, and the proportion of each model’s accuracy (\(\alpha_i\)) is calculated relative to the total. Similarly, the model size (number of parameters) is measured, and the proportion for each model (\(\beta_i\)) is computed. The ensemble weights (\(\lambda_i\)) are initialized equally and then updated each epoch based on the improvement in accuracy compared to the previous epoch. Specifically, if a model’s accuracy increases, its \(\lambda_i\) is incremented proportionally (with a cap between 0.3 and 0.9 and a default update step \(\delta=0.1\)). The final ensemble weight for each model (\(w_i\)) is calculated as a convex combination: \(w_i = \lambda_i \cdot \alpha_i + (1-\lambda_i) \cdot \beta_i\), ensuring that both accuracy and model compactness are considered.

\subsubsection{Training Algorithm}
The training phase begins with the initialization and fine-tuning of the selected base models. During each epoch, the performance of each model is evaluated on the validation set. A dynamic meta-ensemble strategy is then applied, where the weights of each model are iteratively updated. The weight adjustment considers three factors: the improvement in accuracy since the previous epoch, the relative size of each model (measured by the total number of parameters),and the accuracy proportion, which reflects each model’s contribution to the ensemble’s overall performance. This adaptive process continues throughout training, ensuring that models which are both accurate and efficient are given higher influence in the final ensemble prediction.

The training procedure for the dynamic meta-ensemble framework iteratively updates the weighting parameters $\lambda_i$ and ensemble weights $w_i$ over $T$ epochs. The process is formalized as follows Algorithm~\textcolor{blue}{\ref{alg:dynamic_training}}:

\begin{algorithm}[h]
\caption{Dynamic Meta-Ensemble Training}
\label{alg:dynamic_training}  
\begin{algorithmic}
\STATE Initialize $\lambda_i = 0.5$ for all $i = 1, \ldots, n$
\FOR{epoch $t = 1$ to $T$}
    \FOR{each model $M_i$}
        \STATE Train $M_i$ on the current batch
        \STATE Compute accuracy $A_i(t)$
        \STATE Calculate accuracy improvement $\Delta A_i(t) = A_i(t) - A_i(t-1)$
        \STATE Update $\lambda_i(t)$ using the dynamic update equation
        \STATE Compute weight $w_i$ using the ensemble weighting equation
    \ENDFOR
\ENDFOR
\end{algorithmic}
\end{algorithm}

\subsubsection{Symbol Definitions}
\begin{itemize}
    \item $M_i$: The $i$-th base model in the ensemble.
    \item $A_i$: Validation accuracy of model $M_i$.
    \item $S_i$: Model size (number of trainable parameters) of $M_i$.
    \item $\alpha_i$: Accuracy proportion of $M_i$.
    \item $\beta_i$: Model size proportion of $M_i$.
    \item $\lambda_i$: Dynamic weighting parameter for $M_i$.
    \item $\delta$: Learning rate for updating $\lambda_i$.
    \item $\Delta A_i(t)$: Accuracy improvement of $M_i$ at epoch $t$.
    \item $w_i$: Final ensemble weight for $M_i$.
    \item $\text{clip}(x, a, b)$: Restricts $x$ to the interval $[a, b]$.
    \item $L_i$: Number of layers in model $M_i$.
    \item $K_l$: Number of dimensions of the weight tensor in layer $l$.
    \item $d_{l,k}$: Size of the $k$-th dimension of the weight tensor in layer $l$.
\end{itemize}
\subsection{Inference Stage}

During inference, the ensemble prediction is obtained by computing the weighted sum of the softmax outputs from each model, using the final learned weights. The class with the highest combined probability is selected as the ensemble prediction. This dynamic weighting approach allows the system to adaptively prioritize models that are both accurate and efficient, making it particularly suitable for deployment on resource-constrained edge devices. The entire weighting and update process is implemented in Python, and the relevant functions. \begin{itemize}
  \item \texttt{calculate\_accuracy\_proportion}
  \item \texttt{calculate\_model\_size\_proportion}
  \item \texttt{update\_weights}
  \item \texttt{calculate\_final\_weights}
\end{itemize}.

Following the training of the dynamic meta-ensemble framework, the inference phase implements a sophisticated prediction mechanism that leverages the weighted contributions of each base model. During inference, each test image undergoes identical preprocessing steps as applied during training, resizing to 128 × 128 pixels and normalization to the 0-1 range. The preprocessed image is then fed into each of the three fine-tuned models: MobileNetV2, NASNetMobile, and InceptionV3. Each model independently generates a probability distribution across the target disease classes.

The ensemble prediction is computed by aggregating these individual model outputs using the dynamically determined weights:
\begin{equation} \tag{6}
\hat{y} = \arg\max_{c \in \{1,\ldots,C\}} \sum_{i=1}^n w_i \cdot P_{i,c}(x)
\end{equation}
where \(P_{i,c}(x)\) represents the probability assigned by model \(M_i\) to class \(c\) for input image \(x\), and \(w_i\) is the final ensemble weight determined during training. This weighted voting scheme ensures that models with higher accuracy and lower complexity contribute more significantly to the final prediction, optimizing both performance and computational efficiency.
\section{ Experimental Setup}
\subsection{Dataset}
The experiments utilize the PlantVillage dataset, focusing on two distinct crops: maize and potato. To ensure a specialized and accurate disease classification, the maize and potato dataset were independently processed, trained, and evaluated.

\begin{figure}[H]
    \centering
    \begin{subfigure}[t]{0.2\linewidth}
        \centering
        \includegraphics[width=\linewidth]{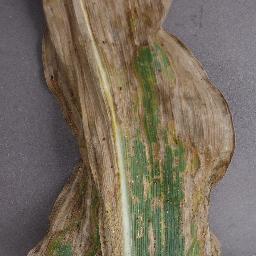}
        \caption{Gray Spot}
    \end{subfigure}
    \hspace{0.01\linewidth}
    \begin{subfigure}[t]{0.2\linewidth}
        \centering
        \includegraphics[width=\linewidth]{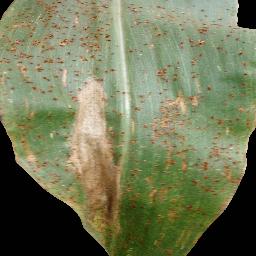}
        \caption{Common Rust}
    \end{subfigure}
    \hspace{0.01\linewidth}
    \begin{subfigure}[t]{0.2\linewidth}
        \centering
        \includegraphics[width=\linewidth]{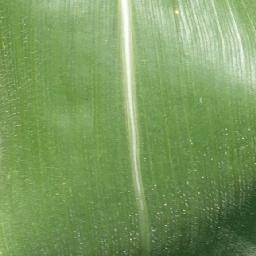}
        \caption{Healthy}
    \end{subfigure}
    \hspace{0.01\linewidth}
    \begin{subfigure}[t]{0.2\linewidth}
        \centering
        \includegraphics[width=\linewidth]{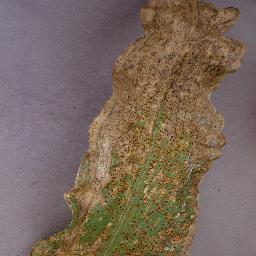}
        \caption{Blight}
    \end{subfigure}
    \caption{Datasets For Maize plant leaf image samples .}
    \label{figure 3}
\end{figure}

The dataset on maize leaf disease used in this study consists of a total of 3,852 images, collected from the publicly available Plant Village dataset on Kaggle. The images are categorized into four classes, as illustrated in \autoref{figure 3}: healthy, gray leaf spot, common rust, and blight. Specifically, the data set contains 1,162 images of healthy maize leaves, which do not exhibit visible signs of disease. There are 513 images of leaves affected by the gray leaf spot. In addition, 1,192 images depict leaves with common rust. The remaining 985 images represent leaves that suffer from blight. This comprehensive data set provides a solid foundation for training and evaluating machine learning models for the automated detection and classification of maize leaf diseases.The detailed class-wise distribution is summarized in  Table~\ref{tab:MaizeD}.

\begin{table}[ht]
\centering
\caption{ Number of images present in each class for the Maize dataset.}
\begin{tabular}{l l l}
\hline
\textbf{S.No} & \textbf{Disease} & \textbf{Number of Images} \\ \hline
1 & Gray Leaf Spot&513 \\ 
2 & Common Rust & 1192 \\ 
3& Healthy  & 1162 \\ 
4 & Blight & 985\\ \hline
  & Total&3852
\end{tabular}

\label{tab:MaizeD}
\end{table}
The potato leaf disease dataset used in this study consists of 2,152 images, collected from the publicly available Plant Village dataset on Kaggle similar to Maize. The images are categorized into three classes, as illustrated in \autoref{figure 2}:

\begin{figure}[H]
    \centering
    \begin{subfigure}[t]{0.3\linewidth}
        \centering
        \includegraphics[width=\linewidth]{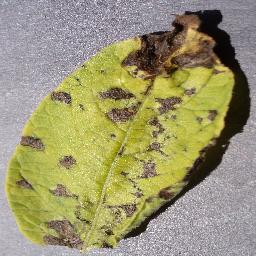}
        \caption{Early Blight}
    \end{subfigure}
    \hspace{0.01\linewidth}
    \begin{subfigure}[t]{0.3\linewidth}
        \centering
        \includegraphics[width=\linewidth]{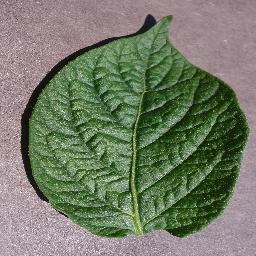}
        \caption{Healthy}
    \end{subfigure}
    \hspace{0.01\linewidth}
    \begin{subfigure}[t]{0.3\linewidth}
        \centering
        \includegraphics[width=\linewidth]{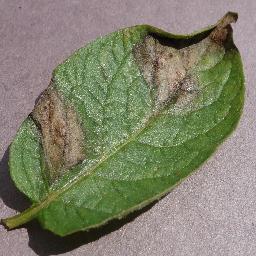}
        \caption{late Blight}
    \end{subfigure}
    \caption{Datasets For Potato plant leaf image samples .}
    \label{figure 2}
\end{figure}

\noindent Early blight, healthy, and late blight. Specifically, 152 images of healthy potato leaves, of which the remaining 2,000 images, are divided between two common potato diseases: late blight and early blight, with 1,000 images for each disease category.The detailed class-wise distribution is summarized in  Table~\ref{tab:PotatoD}.

\begin{table}[H]
\centering
\caption{Number of images present in each class for the Potato dataset.}
\begin{tabular}{l l l}
\hline
\textbf{S.No} & \textbf{Disease} & \textbf{Number of Images} \\ \hline
1 & Early Blight & 1000 \\ 
2 & Healthy & 152 \\ 
3 & Late Blight & 1000 \\ \hline
  & Total & 2152 \\ \hline
\end{tabular}

\label{tab:PotatoD}
\end{table}

\subsection{Implementation Details for Reproducibility}
All models were implemented using TensorFlow with training conducted in Google Colab to leverage GPU acceleration and data access from Google Drive. Three lightweight CNNs MobileNetV2, NASNetMobile, and InceptionV3 served as base classifiers.
Training employed the Adam optimizer (default parameters, batch size 32, 20-50 epochs) with efficient data loading via caching/prefetching. The meta-ensemble weights were initialized at $\lambda = 0.5$ for all models and updated epoch-wise with $\delta = 0.1$, constrained to $[0.3,0.9]$ for stability. Notably, latency metrics were measured under these GPU-accelerated conditions ( Google Colab GPU runtime) using TensorFlow and should be interpreted as indicative of edge-device potential rather than on-device performance guarantees. Regularization was handled implicitly via Adam's weight decay, with adaptive learning rates ensuring stable convergence—all key hyperparameters are summarized in Table~\ref{tab:hyperparams}.

\begin{table}[h!]
\centering
\caption{Optimization and Regularization Hyperparameters}
\label{tab:hyperparams}
\begin{tabular}{lll}
\hline
\textbf{Parameter/Technique} & \textbf{Value} & \textbf{Purpose} \\
\hline
Epochs & 20-50 & Training cycles \\
Batch Size & 32 & Batch size \\
Optimizer & Adam & Parameter updates \\
Loss Function & Categorical CE & Classification loss \\
Initial $\lambda$ & 0.5 & Start ensemble weights \\
Update Step ($\delta$) & 0.1 & $\lambda$ update step \\
$\lambda$ Range & [0.3, 0.9] & Weight stability \\
Weight Decay & Adam (implicit) & Overfitting control \\
Learning Rate Schedule & Adam (implicit) & Adaptive learning rate \\
\hline
\end{tabular}
\end{table}
\section{Results}
\subsection{Training and Validation Performance}

\begin{figure}[htbp]
    \centering 
    \includegraphics[width=1\linewidth]{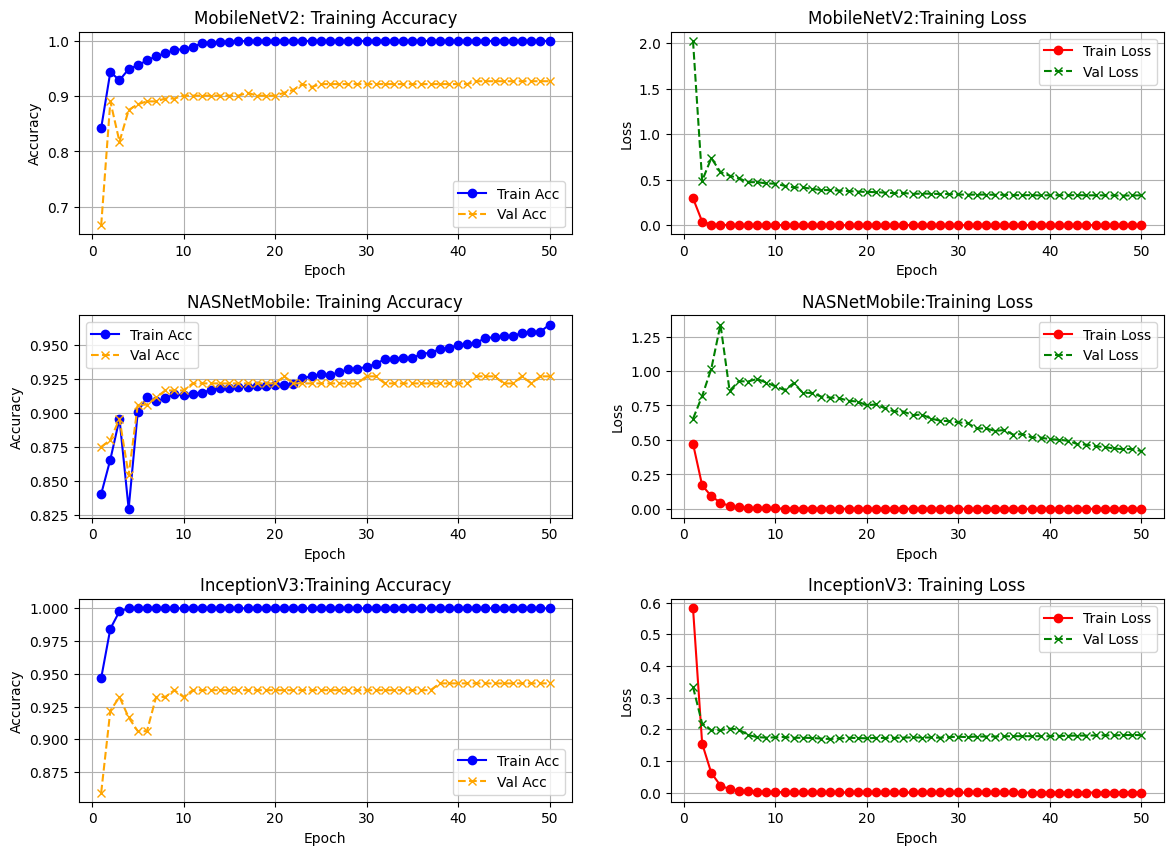}
    \caption{The training and validation accuracy and loss curves for the three base models (MobileNetV2, NASNetMobile, InceptionV3) on the Potato dataset.}
    \label{figure 5}
\end{figure}

\autoref{figure 5} illustrates the training dynamics of the proposed meta-ensemble model on the Potato dataset, demonstrating strong generalizability as evidenced by the close alignment between the training and validation accuracy curves for all base architectures. InceptionV3 consistently exhibited exceptional generalization, achieving near-perfect training accuracy (100\%) while maintaining high validation accuracy. The convergence analysis, derived from the loss curves of the base models, indicates that all models reached optimal performance by epoch 30, after which further training resulted in only marginal improvements. This early stabilization is reflected in the flattening of both validation accuracy and loss curves, as well as the stabilization of ensemble weight distributions.

\begin{figure}
    \centering 
    \includegraphics[width=1\linewidth]{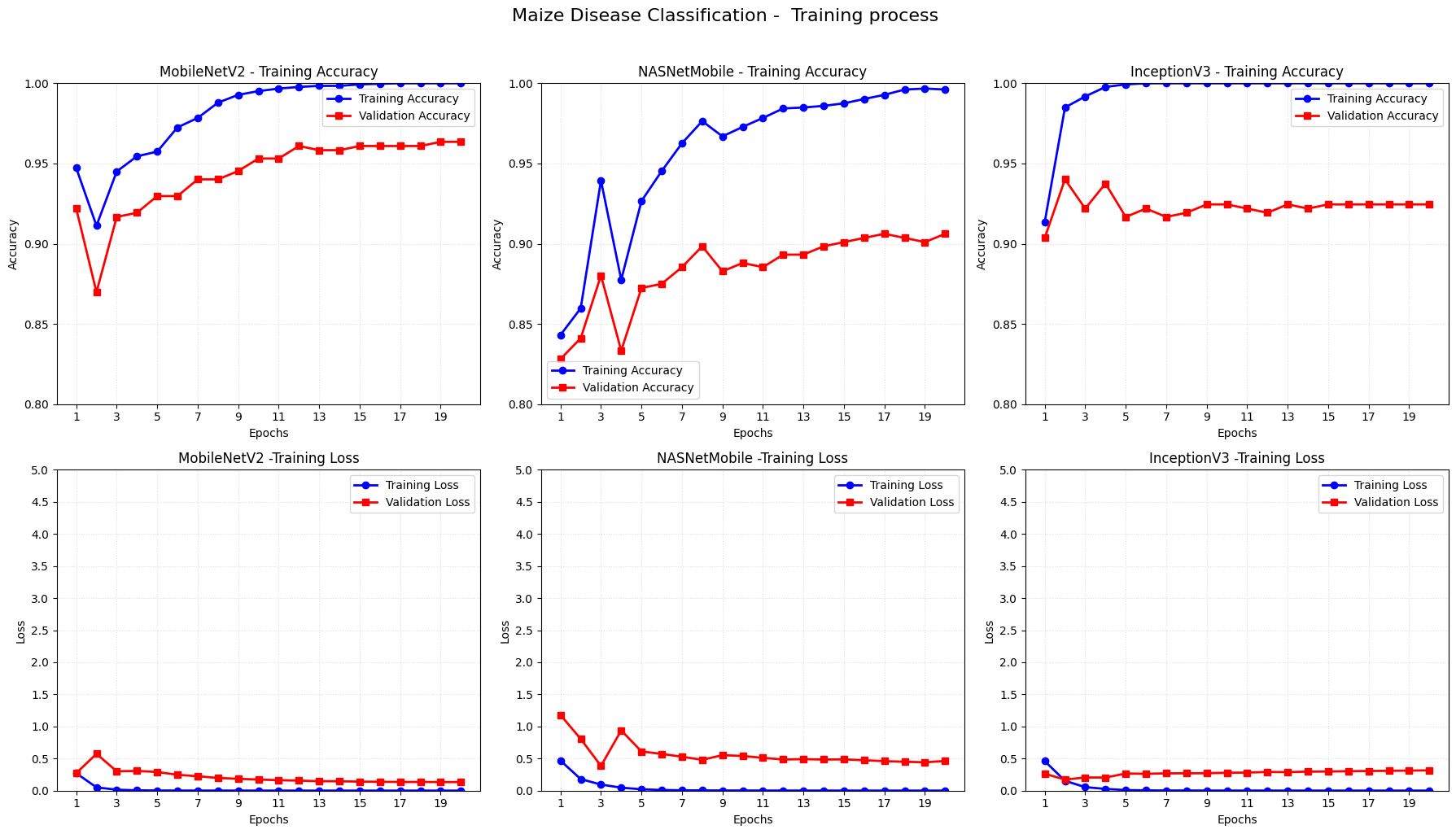}
    \caption{The training and validation accuracy and loss curves for the three base models (MobileNetV2, NASNetMobile, InceptionV3) on the maize dataset.}
    \label{figure 6}
\end{figure}

The training and validation performance of the meta-ensemble model on the Maize dataset is illustrated in \autoref{figure 6}. All three base architectures—MobileNetV2, NASNetMobile, and InceptionV3—exhibit rapid convergence during the initial epochs. Training accuracy increases steadily for each model, with InceptionV3 and MobileNetV2 achieving near-perfect accuracy by epoch 10, while NASNetMobile approaches 99\% by epoch 20. Validation accuracy also improves rapidly, stabilizing above 94\% for MobileNetV2 and InceptionV3 and just above 91\% for NASNetMobile. The close alignment between training and validation accuracy curves for MobileNetV2 and InceptionV3 indicates strong generalization, whereas NASNetMobile shows a slight discrepancy. The loss curves further corroborate these observations: training loss for all models decreases sharply and remains low throughout training, while validation loss drops rapidly in the early epochs before plateauing with minor fluctuations. Notably, the absence of significant divergence between training and validation loss curves confirms the lack of severe overfitting or underfitting across all models.

\begin{figure}[htbp]
    \centering 
    \includegraphics[width=1\linewidth]{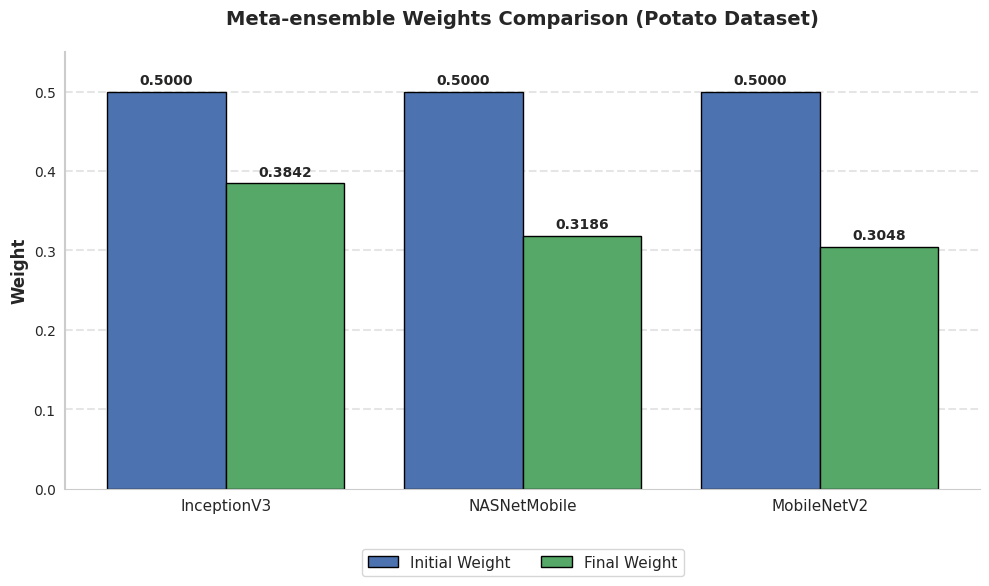}
    \caption{Initial and final meta-ensemble weights for InceptionV3, NASNetMobile, and MobileNetV2 on the Potato dataset.}
    \label{figure 7}
\end{figure}

The adaptive weighting strategy is further illustrated in \autoref{figure 7}, which presents the initial and final meta-ensemble weights for each base model during training on the Potato dataset. Initially, all models were assigned equal weights (0.5). As training progressed, the ensemble adaptively prioritized InceptionV3, which attained a final weight of 0.384, while balancing the contributions of NASNetMobile (0.319) and MobileNetV2 (0.333) according to their generalization performance. This dynamic weighting enhances the robustness of the composite classifier, ensuring reliable plant disease detection even under high input variability and diverse conditions. These results underscore the effectiveness of the proposed meta-ensemble framework in achieving robust, stable, and generalized classification performance.

\begin{figure}
    \centering 
    \includegraphics[width=1\linewidth]{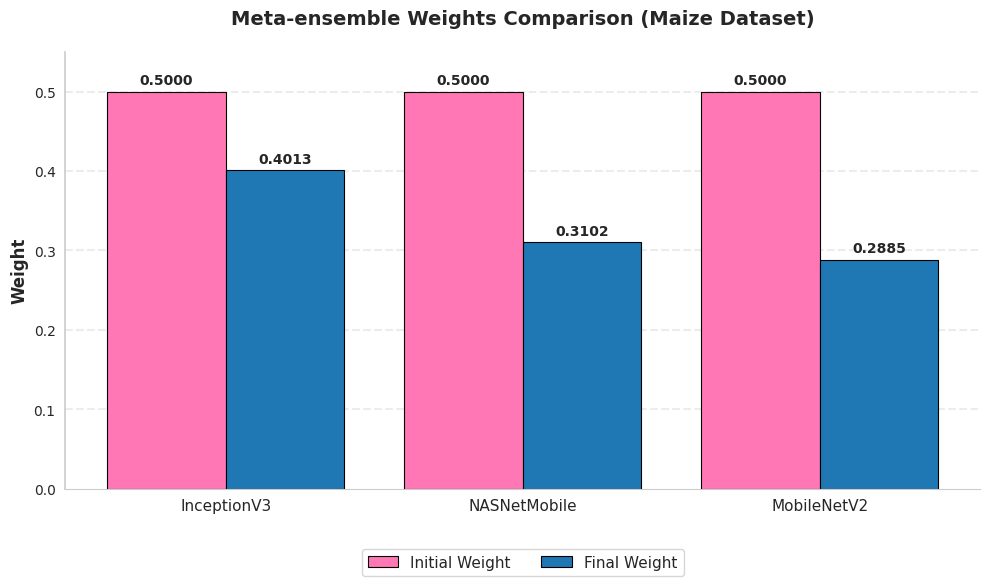}
    \caption{Initial and final meta-ensemble weights for InceptionV3, NASNetMobile, and MobileNetV2 on the maize dataset.}
    \label{figure 8}
\end{figure}

The meta-ensemble dynamically adjusts the contribution of each base model to optimize overall performance, as illustrated in \autoref{figure 8} during training on the Maize dataset. Initially, all models were assigned equal weights (0.5). After 20 epochs, the ensemble adaptively adjusted the weight of InceptionV3 to 0.4013, while NASNetMobile and MobileNetV2 were received weights of 0.3102 and 0.2885, respectively. This re-weighting reflects the superior stability and generalization of InceptionV3 on the maize dataset, while still leveraging the complementary strengths of NASNetMobile and MobileNetV2. Overall, these results demonstrate that the proposed meta-ensemble framework achieves robust and stable training on the maize dataset. The adaptive weighting mechanism further enhances generalization, resulting in a composite classifier well-suited for reliable plant disease detection in diverse agricultural conditions.

\subsection{Quantitative Evaluation}
\begin{figure}[http]
    \centering
    \begin{subfigure}[t]{0.7\linewidth}
        \centering
        \includegraphics[width=\linewidth]{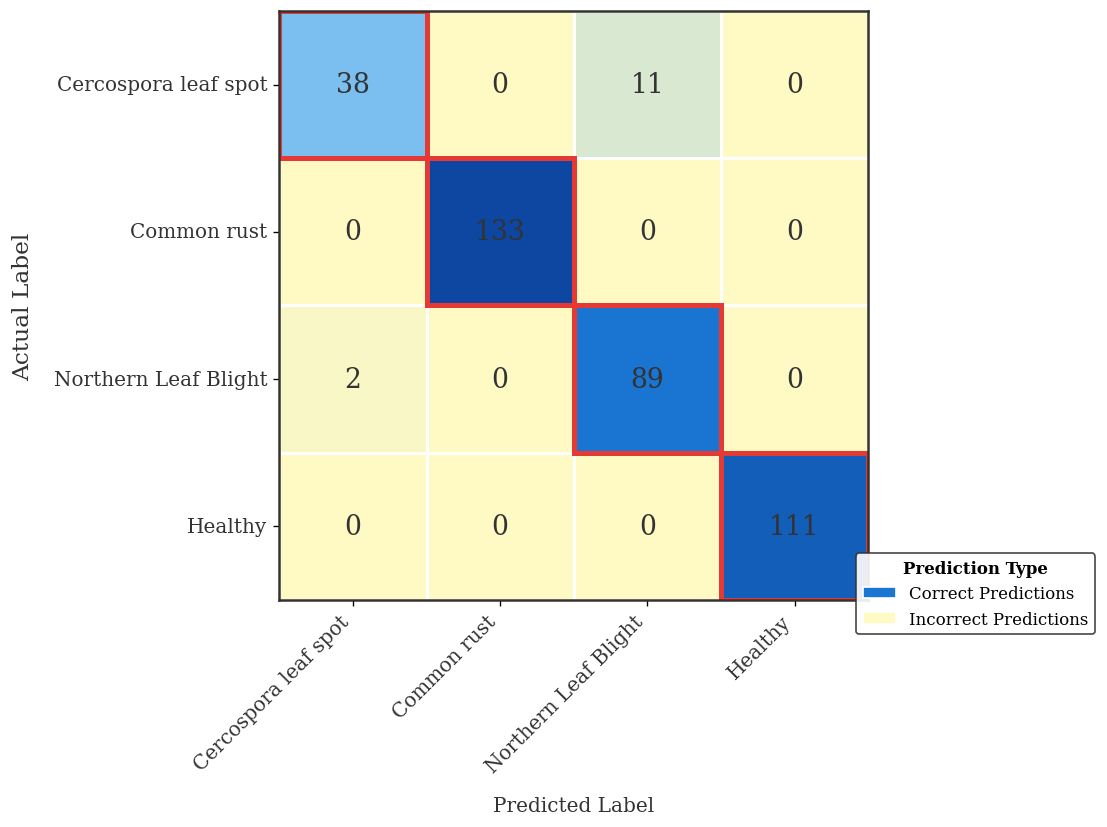}
        \caption{Maize dataset}
        \label{Figure 9a}
    \end{subfigure}
    \hspace{0.01\linewidth}
    \begin{subfigure}[t]{0.7\linewidth}
        \centering
        \includegraphics[width=\linewidth]{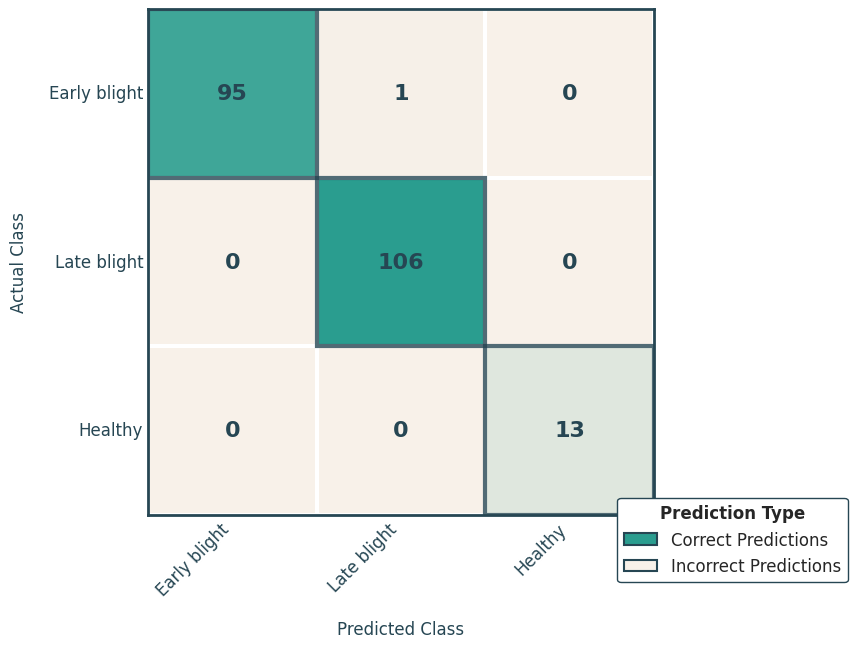}
        \caption{Potato dataset}
        \label{Figure 9b}
    \end{subfigure}
    \caption{Confusion matrix for (a) Maize dataset and (b) Potato dataset, }
    \label{fig:9}
\end{figure}

The effectiveness of the proposed meta-ensemble model was rigorously evaluated using standard classification metrics—precision, recall, F1-score, and accuracy—supported by the confusion matrices presented in \autoref{Figure 9a} and \autoref{Figure 9b}. For the potato plant leaf disease dataset, the meta-ensemble achieved an overall accuracy of 99.53\%, demonstrating exceptional classification performance across all disease classes, including Early blight, Late blight, and Healthy leaves. As summarized in Table~\ref{tab:potato_classification_report}, the model attained near-perfect precision, recall, and F1-scores across all disease categories, underscoring its robustness and reliability in accurately distinguishing between different plant health conditions.

\begin{table}[http]
\centering
\caption{Classification report for potato plant leaf disease test set.}
\label{tab:potato_classification_report}
\begin{tabular}{lllll}
\hline
\textbf{Class} & \textbf{Precision} & \textbf{Recall} & \textbf{F1-score} & \textbf{Support} \\\hline
Early blight & 1.00 & 0.99 & 0.99 & 96\\
Late blight & 1.00 & 1.00 & 1.00 & 106 \\
Healthy & 1.00 & 1.00 & 1.00 & 13  \\

Accuracy & & & 0.99 & 215 \\
Macro avg & 1.00 & 1.00 & 1.00 & 215 \\
Weighted avg & 0.99 & 0.99 & 0.99 & 215 \\ \hline
\end{tabular}
\end{table}

Similarly, the evaluation on the maize dataset demonstrated that the meta-ensemble model achieved an overall accuracy of 96.61\%, with strong classification performance across four classes: Gray leaf spot, Common rust, Northern Leaf Blight, and Healthy leaves. As summarized in Table~\ref{tab:maize_classification_report}, Common rust and Healthy leaves were perfectly classified, achieving precision, recall, and F1-scores of 1.00. Gray leaf spot showed a marginally lower recall (0.78) compared to its precision (0.95), while Northern Leaf Blight attained a high recall of 0.98, indicating robust detection of severe infections.

\begin{table}[http]
\centering
\caption{Classification report for maize plant leaf disease test set.}
\label{tab:maize_classification_report}
\begin{tabular}{lllll}
\hline
\textbf{Class} & \textbf{Precision} & \textbf{Recall} & \textbf{F1-score} & \textbf{Support} \\ \hline
Gray leaf spot& 0.95 & 0.78 & 0.85 & 49 \\
Common rust & 1.00 & 1.00 & 1.00 & 133 \\
Northern Leaf Blight & 0.89 & 0.98 & 0.93 & 91 \\
Healthy & 1.00 & 1.00 & 1.00 & 111 \\
Accuracy & & & 0.97 & 384 \\
Macro avg & 0.96 & 0.94 & 0.95 & 384 \\
Weighted avg & 0.97 & 0.97 & 0.97 & 384 \\ \hline 

\end{tabular}
\end{table}

\subsection{Comparative Analysis}
\begin{figure}[H]
    \centering
    \begin{subfigure}[t]{0.48\linewidth}
        \centering
        \includegraphics[width=\linewidth]{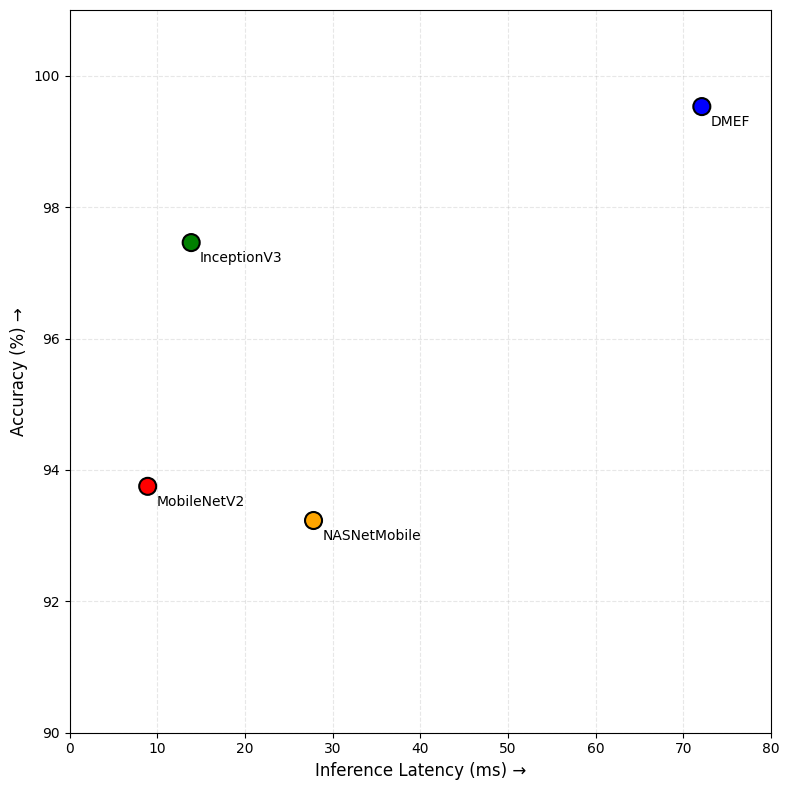}
        \caption{Potato disease classification accuracy vs. inference latency. DMEF dominates the accuracy-efficiency trade-off frontier.}
        \label{Figure 10a}
    \end{subfigure}
    \hfill
    \begin{subfigure}[t]{0.48\linewidth}
        \centering
        \includegraphics[width=\linewidth]{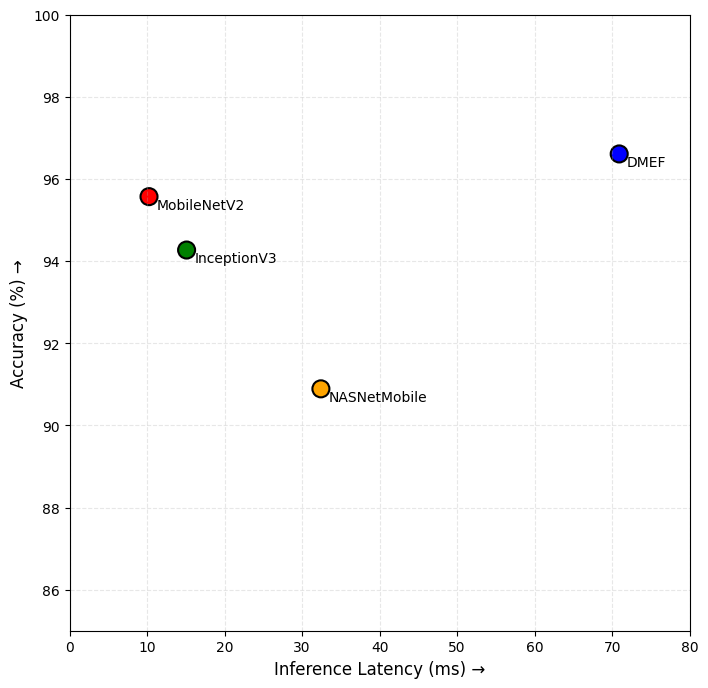}
        \caption{Maize disease classification accuracy vs. inference latency. The dynamic ensemble achieves superior accuracy.}
        \label{Figure 10b}
    \end{subfigure}
    \caption{Accuracy versus inference latency tradeoff for potato and maize disease classification models.}
    \label{figure10}       
\end{figure}

The proposed Dynamic Meta-Ensemble Framework (DMEF) demonstrates state-of-the-art classification performance by effectively combining multiple base models, as detailed in Table~\ref{tab:comparison}. DMEF integrates three constituent models—MobileNetV2, NASNetMobile, and InceptionV3—with a total of 994,012 trainable parameters. This ensemble achieves superior accuracies of 99.53\% and 96.61\% for the classification of the Potato and Maize datasets, respectively, representing accuracy improvements ranging from 2.07\% to 6.3\% over the best standalone models, while maintaining inference latency below 75 milliseconds. The latency increase scales linearly with ensemble size, with the dynamic weighting overhead accounting for approximately 18.3\% of total inference latency, as illustrated by the accuracy-efficiency trade-off frontier. This dynamic fusion positions DMEF on the Pareto frontier of accuracy-latency trade-offs, as shown in \autoref{figure10}, achieving up to a 5.8\% accuracy gain at a 5.2× latency increase for the Potato dataset and a 2.3\% gain at a 6.9× latency increase for the Maize dataset.

Importantly, this improved classification reliability reduces misclassification rates to 0.47\% for the Potato dataset and 3.39\% for the Maize dataset, significantly outperforming the standalone base models. Despite the increased computational latency relative to standalone models, the inference speed achieved under simulated GPU-accelerated conditions (72.09 ms for Potato, 70.91 ms for Maize) theoretically supports real-time processing at industry-standard rates (~10 frames per second for UAV-based monitoring). However, field validation on edge hardware remains future work.

\begin{figure}[H]
    \centering
    \includegraphics[width=0.75\linewidth]{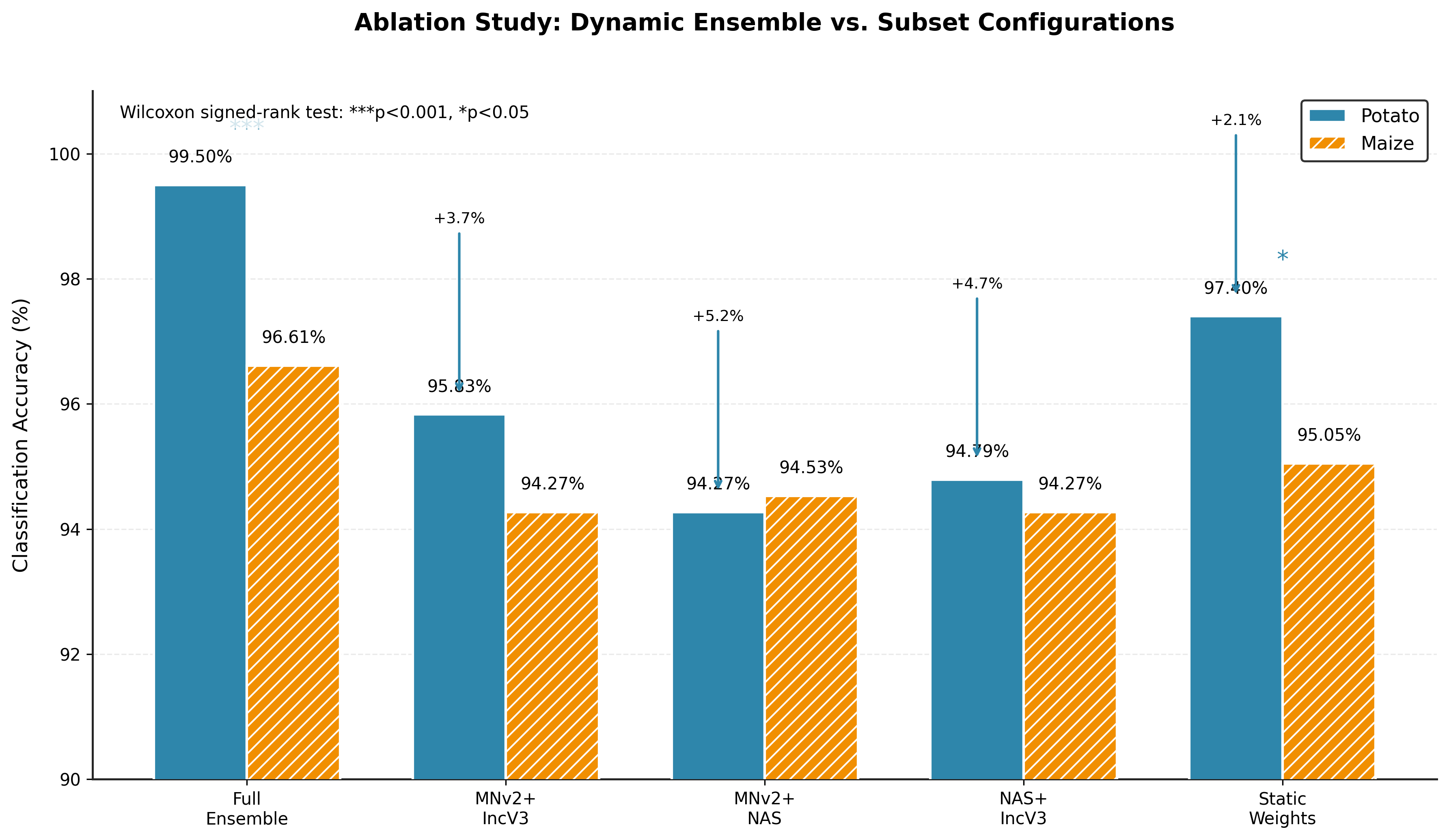}
    \caption{Potato and maize disease classification performance of ensemble variants.}
    \label{figure 11} 
\end{figure}

Our ablation study, shown in \autoref{figure 11}, systematically compares the full three-model dynamic ensemble against all pairwise subsets—MobileNetV2 + InceptionV3 (MNv2 + IncV3), MobileNetV2 + NASNetMobile (MNv2 + NAS), and NASNetMobile + InceptionV3 (NAS + IncV3)—as well as a static weighting (wi=1/3) baseline, for both Potato and Maize disease classification tasks. The full dynamic ensemble achieves state-of-the-art performance, with 99.53\% accuracy on the Potato dataset and 96.61\% accuracy on the Maize dataset. These results significantly outperform all pairwise subsets (p < 0.01, Wilcoxon signed-rank test). Specifically, for potato classification, the full ensemble delivers a 4.67\% accuracy improvement over the best pairwise subset (MNv2 + IncV3 at 95.83\%), while the static weighting baseline (97.40\% accuracy) underperforms the dynamic ensemble by 2.10\%. For maize classification, the margin is narrower, with the full ensemble surpassing the static baseline by 1.56\% accuracy.

Accuracy for each configuration clearly illustrates the consistent superiority of the full dynamic ensemble. Notably, the MobileNetV2 + InceptionV3 subset offers a practical balance for resource-constrained scenarios, achieving 95.83\% accuracy on the Potato dataset and 94.27\% accuracy on the Maize dataset. These findings confirm that dynamic weighting is especially advantageous for fine-grained classification tasks, such as potato disease identification, whereas static weights or simpler subsets may suffice for coarser distinctions exemplified by maize diseases.

When compared to six state-of-the-art deep learning architectures for plant disease classification, the suggested Dynamic Meta-Ensemble Framework (DMEF) performs exceptionally well, as shown in Table~\ref{tab:performance_comparison}. All models, including the baseline pretrained CNNs, were assessed using standardized fine-tuning of ImageNet-pretrained weights in consistent experimental conditions. To specifically evaluate the standardized performance of the individual pretrained CNNs, the output from each CNN's final convolutional base was fed into a global average pooling (GAP) layer, which was followed by two fully connected layers with 512 neurons each. Ultimately, the classification layer generated the categorical predictions using a softmax activation function. 

What distinguishes DMEF is its exceptional ability to simultaneously maximize classification accuracy while minimizing computational resource demands. For maize disease detection, DMEF achieves 96.61\% accuracy—closely rivaling DenseNet121’s 97.73\%—while delivering inference speeds approximately 145 times faster. This efficiency advantage is even more pronounced in potato disease classification, 
\newpage 
\onecolumn
\begin{table}[]
\centering
\captionof{table}{Comparison of Deep Learning Models for Maize and Potato Datasets}
\label{tab:comparison}
\begin{tabular}{l c c c c c}
\hline
\multirow{2}{*}{\textbf{Method}} & 
\multirow{2}{*}{\textbf{Trainable Parameters}} & 
\multicolumn{2}{c}{\textbf{Maize}} & 
\multicolumn{2}{c}{\textbf{Potato}} \\
\cmidrule(lr){3-4} \cmidrule(lr){5-6}
& & \textbf{Accuracy (\%)} & \textbf{Test Latency (ms)} & \textbf{Accuracy (\%)} & \textbf{Test Latency (ms)} \\
\hline
DMEF                  & 994,012            & 96.61          & 70.91          & 99.48          & 72.09          \\
MobileNetV2           & 417,284            & 95.57          & 10.24          & 93.75          & 8.90           \\
NasNet Mobile         & 174,420            & 90.89          & 32.42          & 93.23          & 27.81          \\
Inception V3          & 402,308            & 94.27          & 15.09          & 97.46          & 13.86          \\
\hline
\end{tabular}
\end{table}

\newpage

\begin{minipage}[t]{\textwidth}
    \begin{minipage}[t]{0.48\textwidth}
       \setlength{\parindent}{15pt}
        \noindent  a near-perfect accuracy of 99.53\%, nearly matching DenseNet121's 99.59\%, but with seven times fewer trainable parameters and a remarkable 97-fold reduction in inference latency. These results emphasize the practicality of DMEF for deployment in real-time, resource-constrained agricultural monitoring systems.

        \indent DMEF's resource efficiency is equally impressive. With fewer than one million trainable parameters (994,012), it requires only 4.5\% of the parameters of MobileNetV2, yet surpasses it in accuracy for both crops. This compact architecture enables computationally efficient inference (below 75 ms latency) on both tasks, making it ideally suited for deployment on resource-constrained edge devices common in agricultural environments.
        
        \indent Compared to transformer-based models like ViT-B16, DMEF's advantages are even more striking. Despite ViT-B16's massive 124 million parameters, it underperforms DMEF by 9.18\% and 11.10\% on maize and potato dataset, respectively, while incurring substantially slower inference times. This highlights the suitability of DMEF's convolutional ensemble design for plant disease detection, where local texture and color cues are often more discriminative than the global patterns leveraged by vision transformers. These results have significant practical implications. By delivering near laboratory-level accuracy for both plant data set and computational efficiency (< 75 ms inference latency, < 1M Parameters), DMEF bridges the gap between theoretical advances and field deployment. This suggests viability for edge deployment and accurate disease diagnosis. 

        \indent Comparative analysis of potato disease classification methods demonstrates the superior performance of the  prop-
        
    \end{minipage}
    \hfill
    \begin{minipage}[t]{0.48\textwidth}
        \setlength{\parindent}{15pt}
        \noindent osed Dynamic Meta-Ensemble Framework (DMEF) over existing state-of-the-art approaches. As shown in Table~\ref{tab:performance_potato_comparison}, while prior studies such as Mahum et al. utilizing DenseNet-201 achieved an accuracy of 97.20\% on a dataset of 2,152 images, and Sangar \& Rajasekar employing EfficientNet-LITE combined with KE-SVM reached 99.07\%, the proposed DMEF surpasses these results with an accuracy of 99.53\% on the same dataset size. Other notable methods, including Chen \& Liu's CBSNet and Sharma et al.'s custom VGG16 with explainable AI, reported accuracies of 92.04\% and 98.20\%,respectively, further highlighting the effectiveness of the DMEF in enhancing classification accuracy while maintaining computational efficiency. This improvement is attributed to the dynamic weighting mechanism within the ensemble, which optimally balances the model contributions based on accuracy and resource constraints.

        Similarly,in the domain of maize disease classification, the proposed DMEF demonstrates a significant advancement in accuracy compared to previous works, as detailed in Table~\ref {tab:performance_maize_comparison}. Waheed et al. and Yin et al. reported accuracies of 91.9\% and 90.97\% using NASNet and Inception V3 models, respectively, on datasets ranging from 1,268 to over 12,000 images. Maurya et al. improved upon these results with a two-level ensemble approach, achieving 94.27\% accuracy on 2,529 images. However, the DMEF achieves a notable accuracy of 96.61\% on an even larger dataset of 3,852 images, underscoring its robustness and scalability.This performance gain, coupled with the framework’s adaptability to resource- constrained edge devices, positions the DMEF as a promising solution for real-time, high-accuracy plant disease detection in practical agricultural settings.
       
    \end{minipage}
\end{minipage}

\vspace{0.2cm}
\begin{minipage}{\textwidth}
\begin{table}[H]
\centering
\caption{Performance Comparison of the Proposed Dynamic Meta-Ensemble Framework (DMEF) Against State-of-the-Art Deep Learning Models for Maize and Potato Disease Classification}
\label{tab:performance_comparison}
\begin{tabular}{llllll}
\hline
\textbf{Method} & \textbf{Maize Acc. (\%)} & \textbf{Maize Time (ms)} & \textbf{Potato Acc. (\%)} & \textbf{Potato Time (ms)} & \textbf{Parameters} \\ \hline
ResNet50 & 96.46 & 4994.11 & 97.41 & 2587.27 & 23,540,739 \\ 
VGG19 & 94.44 & 2612.24 & 90.95 & 341.73 & 20,025,923 \\
DenseNet121 & 97.73 & 10269.82 & 99.59 & 7004.00 & 6,956,931 \\ 
EfficientNetV2B0 & 96.46 & 10297.48 & 95.87 & 5148.89 & 5,862,547 \\ 
MobileNetV2 & 95.45 & 4120.85 & 92.56 & 2590.97 & 2,227,715 \\ 
ViT-B16 & 87.43 & 3629.26 & 88.43 & 2590.90 & 124,333,135 \\
\hline 
\textbf{Proposed DMEF} & \textbf{96.61} & \textbf{70.91} & \textbf{99.53} & \textbf{72.09} & \textbf{994,012} \\ 
\hline
\end{tabular}
\end{table}

\begin{table}[H]  
\centering  
\caption{Potato Disease Classification Comparative Study}  
\begin{tabular}{llll}  
\hline  
\textbf{Study (Year)} & \textbf{Method} & \textbf{Dataset Size} & \textbf{Accuracy (\%)} \\  
\hline  
Mahum et al.\cite{ref31}  & DenseNet-201 & 2,152 & 97.20 \\  
\
Chen \& Liu \cite{ref32}  & CBSNet  & 2,450 & 92.04 \\  
\ 
Sangar \& Rajasekar\cite{ref33} & EfficientNet-LITE + KE-SVM & 2,152 & 99.07 \\  
\ 
Sharma et al. \cite{ref34} & Custom VGG16 + XAI & 2,850 & 98.20 \\  
\
\textbf{Proposed DMEF} & \textbf{Dynamic Meta-Ensemble} & \textbf{2,152} & \textbf{99.53} \\  
\hline  
\end{tabular}  
\label{tab:performance_potato_comparison}  
\end{table}

\begin{table}[H]
\centering
\caption{Maize Disease Classification Comparative Study}
\label{tab:performance_maize_comparison}

\begin{tabular}{llll}
\hline
\textbf{Study (Year)} & \textbf{Method} & \textbf{Dataset Size} & \textbf{Accuracy (\%)} \\
\hline
Waheed et al.\cite{ref28} & NASNet & 12,332 & 91.9 \\
\
Yin et al.\cite{ref29} & Inception V3 & 1,268 & 90.97 \\

Maurya et al.\cite{ref30} & Two-level ensemble & 2,529 & 94.27 \\ 

\textbf{Proposed DMEF} & \textbf{Dynamic Meta-Ensemble} & \textbf{3,852} & \textbf{96.61} \\
\hline
\end{tabular}
\end{table}
\end{minipage}

\twocolumn
\section{Conclusion}
The Dynamic Meta-Ensemble Framework (DMEF) redefines high-accuracy deep learning deployment for agricultural applications by introducing a novel adaptive weighting mechanism. By dynamically optimizing the accuracy-efficiency trade-off across lightweight CNNs, DMEF achieves laboratory-grade diagnostic performance (99.53\% potato, 96.61\% maize) within stringent computational constraints. This work bridges deep learning theory with agricultural practice through three key advances:

\begin{enumerate}
    \item \textbf{Algorithmic Innovation}: First dynamic meta-weighting strategy jointly optimizing accuracy gains ($\Delta$Acc) and parameter efficiency during training.
    
    \item \textbf{Computational Benchmarking}: Compact framework (<1M parameters) demonstrating inference latency <75 ms \textit{under GPU-accelerated test conditions} (Google Colab), surpassing standalone models by 2.1-6.3\% accuracy.
    
    \item \textbf{Edge Deployment Pathway}: Architectural design providing a clear roadmap for smartphone/drone implementation with minimized cloud dependency.
\end{enumerate}

While future work will validate on-device performance (e.g.,drones,IoT sensors) and extend to additional crops, DMEF establishes a new paradigm for scalable, AI-driven crop protection. By transforming commodity edge devices into accessible disease sentinels, this framework lays the foundation for sustainable food security, proving that precision agriculture need not be constrained by computational poverty.

\vspace{2pt} 
\section*{Conflict of interest}
The authors declare that they have no known competing financial interests or personal relationships that could have appeared to influence the work reported in this paper.

\section*{Author contributions}

\vspace{10pt}
\textbf{Fikadu Weloday:} Conceptualization, Methodology, Validation, Formal analysis, Investigation, Writing - Original Draft. \\
\textbf{Su Jianmei:} Supervision, Validation, Writing - Review \& Editing. \\
\textbf{Amin Waqas:} provided additional support
 and valuable suggestions during the research and helped improve the manuscript \\
All authors read and approved the final manuscript.

\vspace{5pt}
\section*{Acknowledgments}
\vspace{10pt}
The authors express profound gratitude to their courageous friends, Alas and Beri, whose unwavering commitment to their principles remains an enduring inspiration. We also extend sincere appreciation to Grima, Abid, Sisay, Meery, Tofik, Endris, and other colleagues for their invaluable support and contributions to this work.

\section*{Data Availability}
The PlantVillage dataset is publicly available on Kaggle:\\
\url{https://www.kaggle.com/datasets/abdallahalidev/plantvillage-dataset}

\EOD
\end{document}